%% file: sn-article.tex
\documentclass[sn-mathphys,Numbered]{sn-jnl}
\usepackage{graphicx}%
\usepackage{multirow}%
\usepackage{amsmath,amssymb,amsfonts}%
\usepackage{amsthm}%
\usepackage{mathrsfs}%
\usepackage[title]{appendix}%
\usepackage{xcolor}%
\usepackage{textcomp}%
\usepackage{manyfoot}%
\usepackage{booktabs}%
\usepackage{algorithm}%
\usepackage{algorithmicx}%
\usepackage{algpseudocode}%
\usepackage{listings}%
\usepackage{caption}
\usepackage{subcaption}

\begin{document}
\title[Article Title]{Virtual Mines – Component-level recycling of
printed circuit boards using deep learning}

\author[1]{\fnm{Muhammad} \sur{Mohsin}}

\author[1,2]{\fnm{Stefano} \sur{Rovetta}}

\author[1,2]{\fnm{Francesco} \sur{Masulli}}

\author[2]{\fnm{Alberto} \sur{Cabri}}


\affil[1]{\orgdiv{DIBRIS, \orgname{University of Genoa}, \country{Italy}}}

\affil[2]{\orgname{Vega Research Laboratories S.R.L.}, \country{Italy}}


\abstract{ This contribution gives an overview of an ongoing project using machine learning and computer vision components for improving the electronic waste recycling process. In circular economy, the “virtual mines” concept refers to production cycles where interesting raw materials are reclaimed in an efficient and cost-effective manner from end-of-life items. In particular, the growth of e-waste, due to the increasingly shorter life cycle of hi-tech goods, is a global problem. In this paper, we describe a pipeline based on deep learning model to recycle printed circuit boards at the component level. A pre-trained YOLOv5 model is used to analyze the results of the locally developed dataset. With a different distribution of class instances, YOLOv5 managed to achieve satisfactory precision and recall, with the ability to optimize with large component instances.}
\maketitle
\section{Introduction}\label{sec1}
Due to the high demand of modern electronic devices, electrical and electronic waste is increasing exponentially. At present, Asia is one of the largest producers of e-waste, while Europe has the highest rate of production per person (16.2 kg/person-year) \cite{forti2020global}. Thus, recycling of these e-waste is very important both economically and environmentally. As compared to urban mining, traditional mining methods are not as cost-effective and environmentally friendly. The concept of urban mining refers to the extraction of valuable materials from e-waste in a cost-effective manner \cite{zeng2018urban}. Due to the physical process of dismantling, recycling all the products and components is not feasible. Therefore, component level recycling of products containing the most valuable materials is a viable option. 

In all e-waste products, printed circuit boards (PCBs) contain the most resources which are crucial to the production of new electronic devices. Waste of electrical and electronic equipment (WEEE) contains high yield materials called critical raw materials (CRMs), which can be reused in new production processes and reduce pollution. Several of these materials are not available within the European Union (EU) and the recycling process is nearly nonexistent. According to the latest European Commission report about CRMs \cite{european2020communication}, 30 materials are declared as critical due to its high demand and dependency on other countries like china and therefore need to be recycled and reused within EU. The EU evaluates the global production and consumption of these materials every three years and adds or removes them from the critical list. A report was published in 2011 \cite{european2011communication} that identified 14 elements as critical, this was followed by 20 elements in 2014 \cite{european2014communication}, 27 elements in 2017 \cite{european2017communication} and 30 elements in 2020 \cite{european2020communication} based on their high demand and unavailability in Europe, those are declared as critical raw materials.

This paper describe the detail system used for component level recycling of waste PCBs using deep learning techniques. Local V-PCB (Vega Research Laboratories) dataset is used to analyse results of selected deep learning model for individual PCB components. Usually, an electronic board is made up of 45-50 chemical elements including rare earth elements. These elements consist of 30 \% metallic and 70 \% non-metallic elements. Some of the elements are heavy metal which can be seriously damaging to the environment and human
health. At present, only 10 to 15 chemicals are partially recycled from WEEE; all others are lost. Details about the electronic components and chemical elements including critical raw materials present in them is shown in Table \ref{tab:table1}:

\begin{table}[h!]
    \centering
    \caption{Electronic board composition in PCBs.}
    \label{tab:table1}
    \begin{tabular}{|c|c|} 
          \hline
      \textbf{Electronic Components} & \textbf{Critical Raw Materials (CRMs)} \\
   
      \hline
     
Capacitors & Ta, Pd, Nb \\
      \hline
Resistors  & Ru, Ta \\
      \hline
Semiconductors  & Ga, Ge, In, Sb, Ta \\
      \hline
Transistors  & Ga, Ge  \\
      \hline
ICs (Capacitors, Resistors,Semiconductors, Transistors)& Contain all above CRMs\\ 

      \hline
Connectors  & Pd, Ru, Be \\
      \hline
Plating  & Ni, Cu, Au \\
\hline      
    \end{tabular}  
\end{table}

\begin{figure}
\centering
 \begin{subfigure}{.5\textwidth}
   \centering
   \includegraphics[width=.85\linewidth]{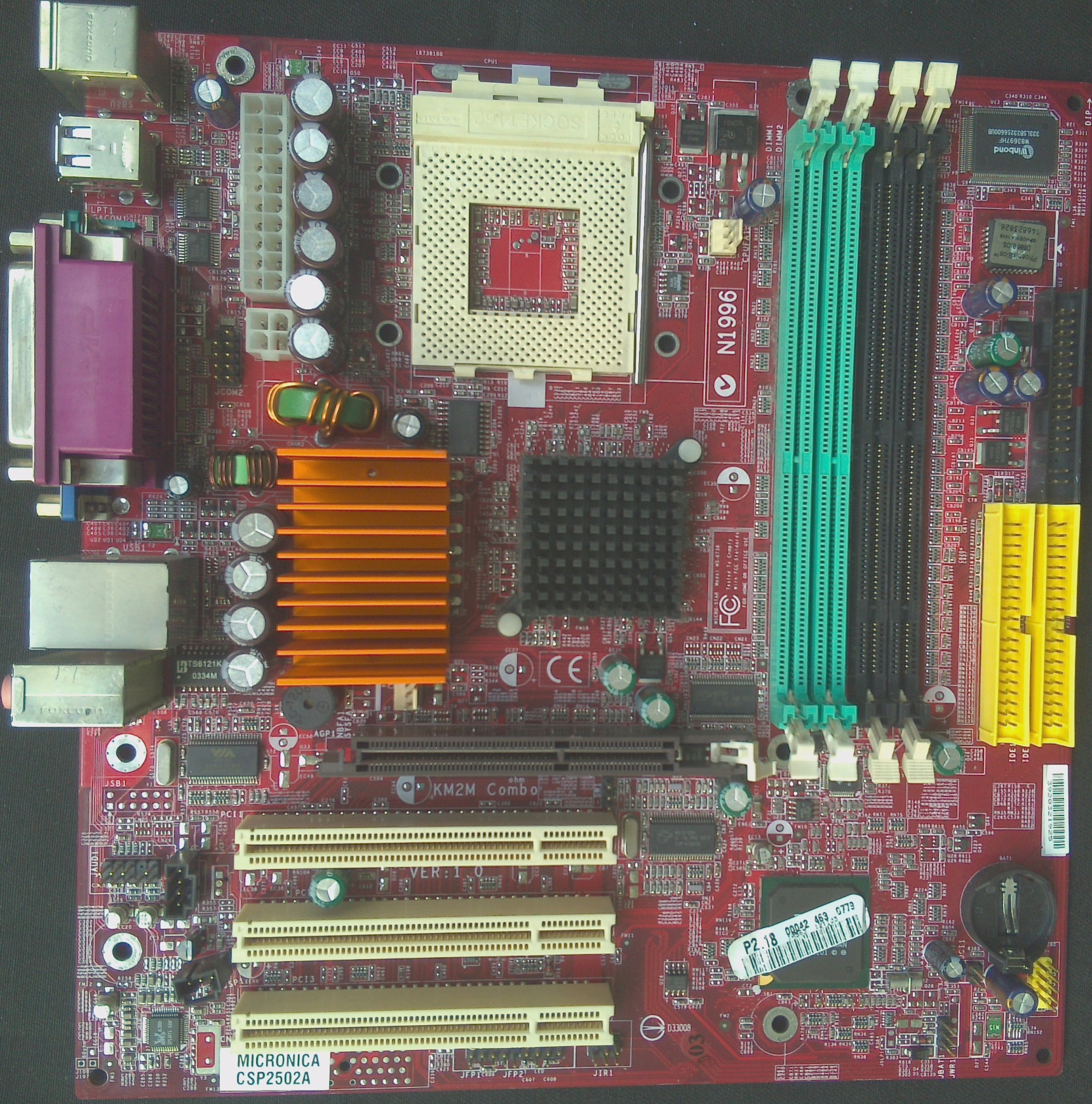}
   \caption{PCB-1}
   \label{fig:sub1}
 \end{subfigure}%
 \begin{subfigure}{.5\textwidth}
   \centering
   \includegraphics[width=.815\linewidth]{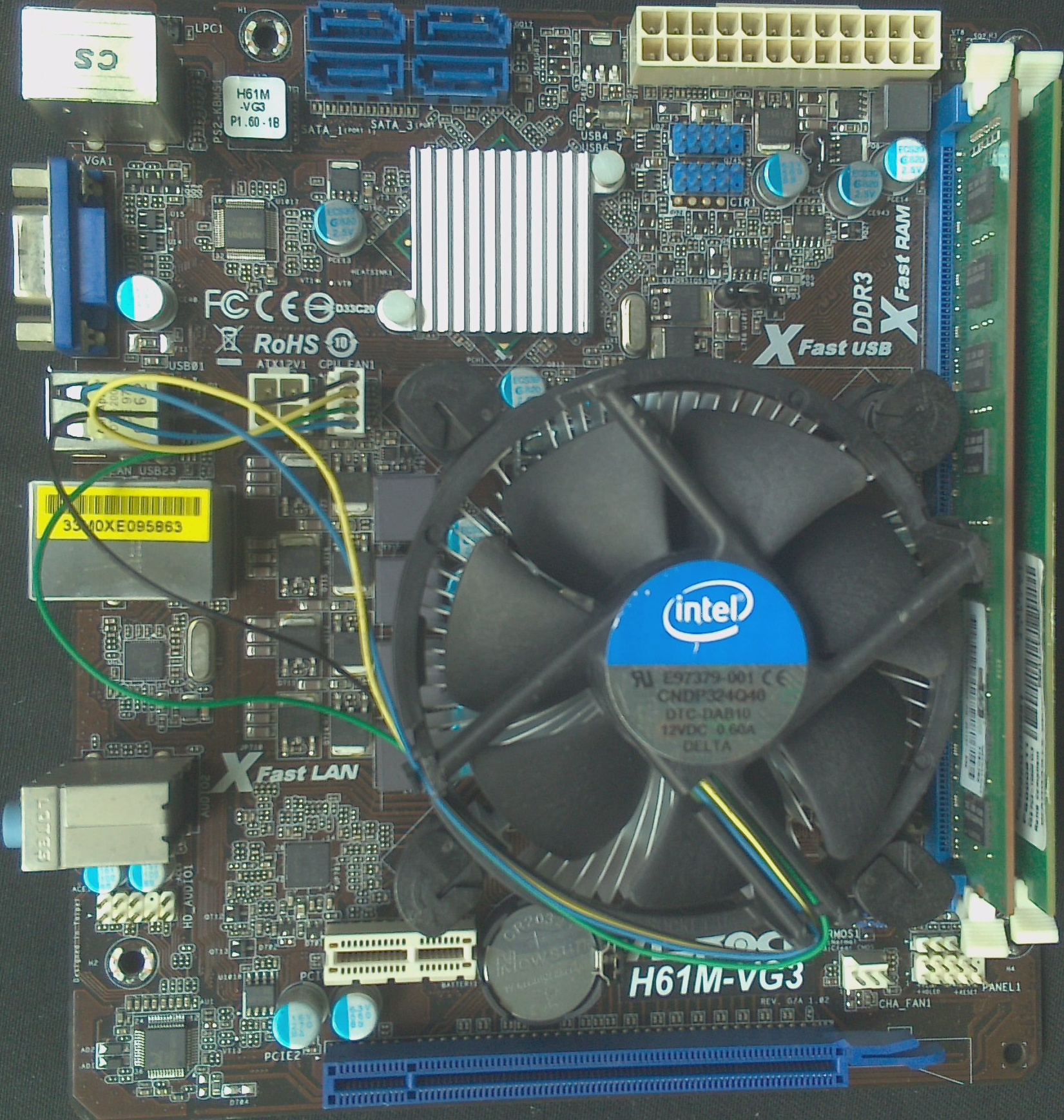}
   \caption{PCB-2}
   \label{fig:sub2}
 \end{subfigure}
\caption{V-PCB dataset example}
\label{fig:test}
\end{figure}
This paper contains five sections. Section 2 gives an overview of the state of the art techniques used for e-waste recycling. 3  methodology, 4 results and discussion, and last section summarizes the results in the form of conclusion. 

\section{Related Work}\label{sec2}
With the emergence of the circular economy, many stakeholders, companies and small businesses have become increasingly interested in extracting valuable resources from waste (e-waste).  Communities are widely discussing the concept of recycling, land fill mining, and securing resources for re-use in new production processes \cite{mining2015urban}. The circular economy model refers to a closed loop system which relies on the strategy of reduce, reuse, and recycle in order to achieve sustainability. By minimizing resource consumption, this model reduces waste and stimulates economic growth while preserving the environment.

Xavier \cite{xavier2021sustainability} discussed that urban mining is a sustainable way to exploit mineral resources that discourage the extraction of minerals through traditional mining, thus reducing environmental impacts and preserving natural resources. Many conventional approaches have been used in the past to extract valuable materials from e-waste, including pyrometallurgical and hydrometallurgical techniques, which are rapid and efficient, but often cause secondary pollution and are not economically sustainable \cite{priya2017comparative}. The process of bioleaching has also been widely used to extract precious metals from ores \cite{xiang2010bioleaching,nguyen2021bioleaching,yaashikaa2022review}. The use of these physical and chemical conventional techniques is associated with many disadvantages, including: time consuming, low efficiency, high capital costs, labour intensive, and environmental concerns \cite{yaashikaa2022review}.

Modern advancements in technology have made it possible to develop many artificial intelligence (AI) applications that support circular economy in recycling activities \cite{cabri2022recovering}. AI now analyzes large quantities of data at high speed and can support CE through the entire value chain, from demand prediction to re-manufacturing \cite{kamble2022digital}. AI based models are mainly data driven and require large amounts of data for training and optimization. This data can be collected from many sources, such as publicly available datasets, user-based data, and sensor data. AI models are then evaluated based on their accuracy and performance \cite{mohsin2021automatic}. The PCB board is the building block of every electronic device, and it contains the highest amount of valuable materials in comparison to other e-waste. Different AI based techniques have been develop for PCB: defect detection \cite{adibhatla2018detecting}, visual inspection \cite{lu2020fics}. Different deep learning based approaches \cite{li2013smd,li2016localizing,li2014text} were proposed for automatic PCB recycling, including sensors, lasers and cameras, segmentation and localization techniques using borders around PCBs, and clustering techniques based on color features for localization. Additionally, he presented IC identification based on OCR.

The paper \cite{xu2020n}, presented a model for recognizing electronic components using deep learning network. As a result of studying different deep learning network models including ResNet, SquezeNet, DenseNet, MobileNetV2, and EfficientNet \cite{huang2016learning,japkowicz2002class,fan2019birnet,tan2019efficientnet}, they proposed an improved SquezNet architecture for PCB component recognition that is both faster and more accurate.

\section{Methodology}\label{sec3}

\begin{figure}[!htb]
\centering
\includegraphics[width=0.90\textwidth]{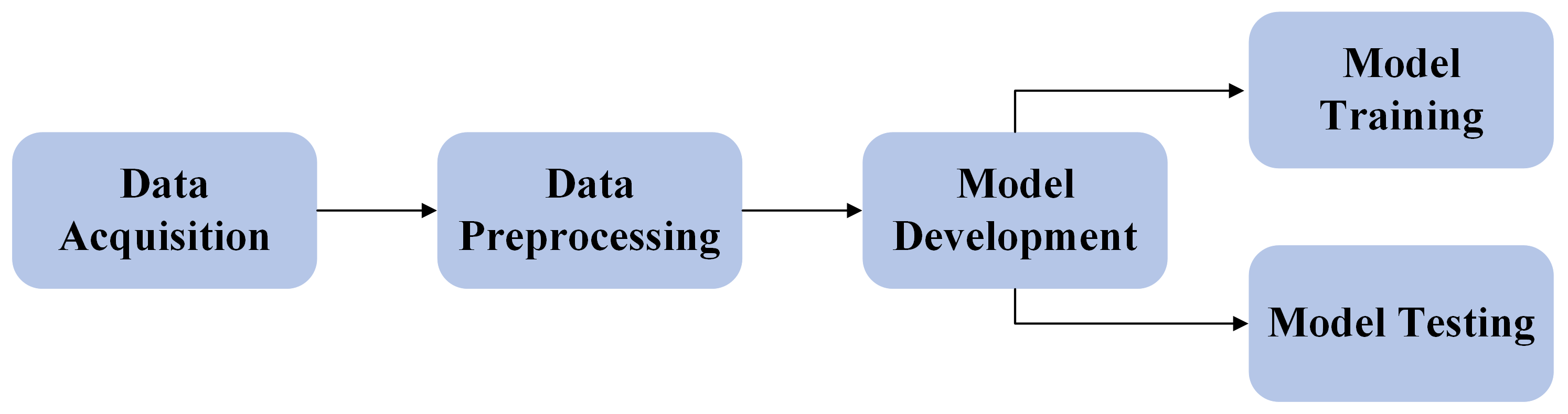}
\caption{Illustration of the component-level recycling workflow of PCBs using deep learning}
\label{fig:image1}
\end{figure}

This section provides an overview of a methodology used to recycle PCBs from e-waste using deep architectures. Figure \ref{fig:image1} shows the detail illustration of working methodology pipeline. It includes the data acquisition, data preprocessing, model development followed by model training and model testing. Each step is explained in the following.

\subsection{Data Acquisition}\label{subsec3-1}
The acquisition of data is one of the most challenging aspects of developing deep learning models. Gathering data is not only time-consuming, but can also be expensive and requires a complete pipeline from the collection, and pre-processing to an annotation. Additionally, the deep learning model will need to be trained on a large amount of data to be accurate, and this requires significant computational resources.
In order to acquire data, several PCB plants are visited and PCB boards are collected, which are then used to create the V-PCB dataset. An environmental setup consists of a Raspberry Pi camera connected to NIVIDIA Jetson Nano. The Jetson Nano is a powerful, small computer designed specifically for edge devices due to its high levels of parallel processing.  For training deep learning model, we  used in total 42 images (30 for training  and 12 for validation) captured with Jetson camera module.
\begin{table}[h!]
\caption{Environmental setup}
\label{tab:table2}
\begin{tabular}{|c|c|}
\hline
\textbf{NVIDIA Jetson Nano} & \textbf{Raspberry Pi Camera V2} \\
\hline
 NVIDIA Maxwell architecture GPU 128 CUDA cores & 8 megapixels  \\
\hline
Quad-core ARM Cortex-A57 CPU clocked at 1.43 GHz & 3280x2464 pixels (highiest) \\
\hline
\end{tabular}
\end{table}
\subsection{Data Preprocessing}\label{subsec3-2}
The V-PCB data contains raw images of PCBs captured using a Raspberry Pi camera connected to a Jetson Nano. In order to analyze these raw image data with computer vision based-approaches, it is necessary to pre-process them into dataset that are appropriate for computer vision-based analysis. This step aims to reduce the noise in V-PCB images and to enhance their quality. V-PCB raw images contain an extensive background area that is neither useful for detection nor computationally efficient. We extracted region of interest (ROI) from V-PCB images and these ROIs contain all of the components, which are then detected one by one using a deep learning model.

\begin{figure}[H]
\centering
\includegraphics[width=0.75\textwidth]{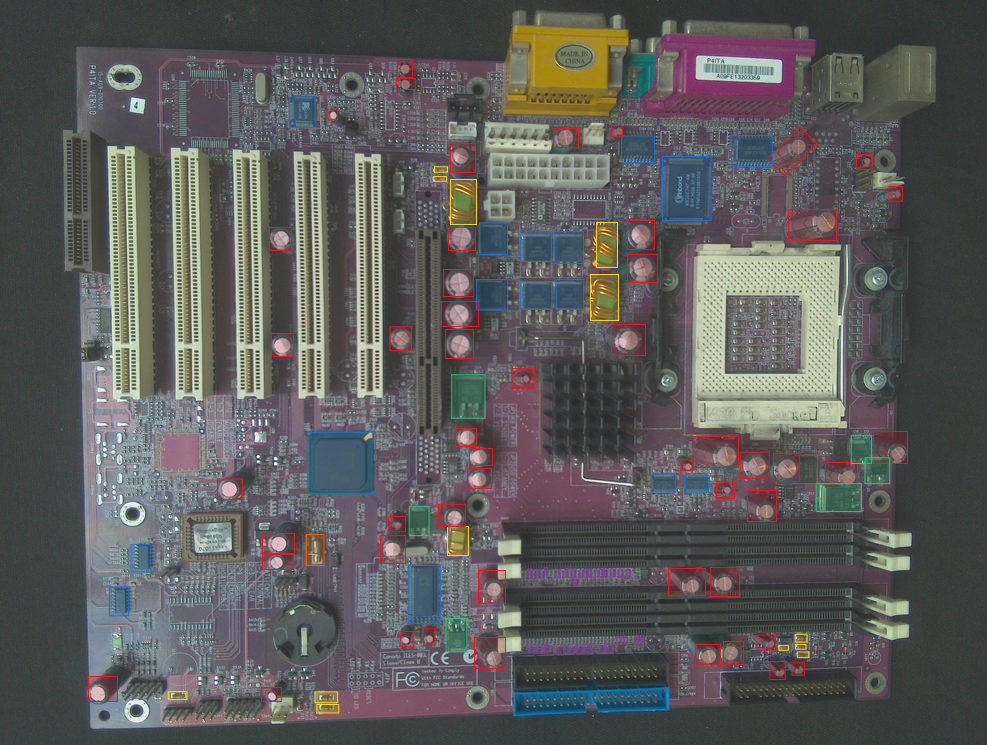}
\caption{Data Labeling using Label Studio; online open source software \cite{labelstudio}}
\label{fig:labeling}
\end{figure}

Data annotation is performed by using online open source tool called label studio \cite{labelstudio} available for object detection applications. There are several types of components that contain critical raw materials on PCBs \cite{european2020communication}. Figure \ref{fig:labeling} shows the annotation representation along with classes.
In this study, we identified the eight components: capacitor, electrolytic capacitor, diode, integrated circuits, transistor, resistor, coil and transformer and considered as class in training pipeline. The selection of PCB components is totally dependent on the quantity of critical raw materials present in them. For example capacitor contains Ta, Pd and Nb.
\vspace{10pt}
\subsection{Model Development and Inference}\label{subsec3-3}
Deep architecture models require large amount of data for computer vision based applications. Therefore, before moved into the step of model design and implementation, V-PCB dataset is prepared. All the PCB images along with the bounding box information as label are given as an input to the model pipeline. 

In general, object detection model pipeline consist of following steps: V-PCB image as input and then performed pre-processing like image enhancement, contrast adjustment and extracting region of interest and then pass it to the object detection model. YOLOv5 pre-trained weights are used for the training and testing the model. YOLOv5 \cite{jocher2022ultralytics} is latest object detection model and used in real time application.The model is visualized in Figure \ref{fig:model}:
\begin{figure}[H]
\centering
\includegraphics[width=1\textwidth]{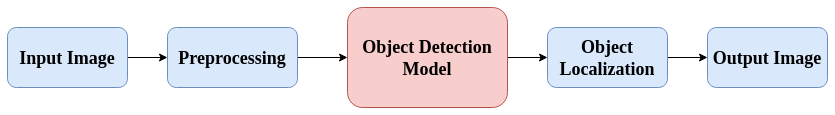}
\caption{Object Detection Model Pipeline}
\label{fig:model}
\end{figure}
Python 3.6 along with openCV is used for all type of image processing tasks such image enhancement, contrast adjustment and extracting region of interest. For training YOLOv5 model, linux server dedicated with high speed GPU cuda supported have used and inference is performed on Jetson Nano environment.

\section{Results and Discussion}\label{sec4}
The study aim is to enhance the collection, disassembly, and recycling processes for PCBs. The project is divided into a number of phases, including creating an image database of PCBs, building a hardware device for visual classification and training a visual classifier for components.

The study covered how important urban mining and the circular economy are to recycling electronic waste. It brought attention to the necessity of effective sorting techniques for recovering valuable materials from e-waste. Conventional methods for e-waste material extraction frequently suffer from drawbacks like poor efficiency, high costs, and environmental concerns. As a result, the use of artificial intelligence (AI) and deep learning models to support circular economy activities, such as PCB recycling, has gained attention. 

\begin{figure}
\centering
\includegraphics[width=0.80\textwidth,height=0.5\textheight,keepaspectratio]{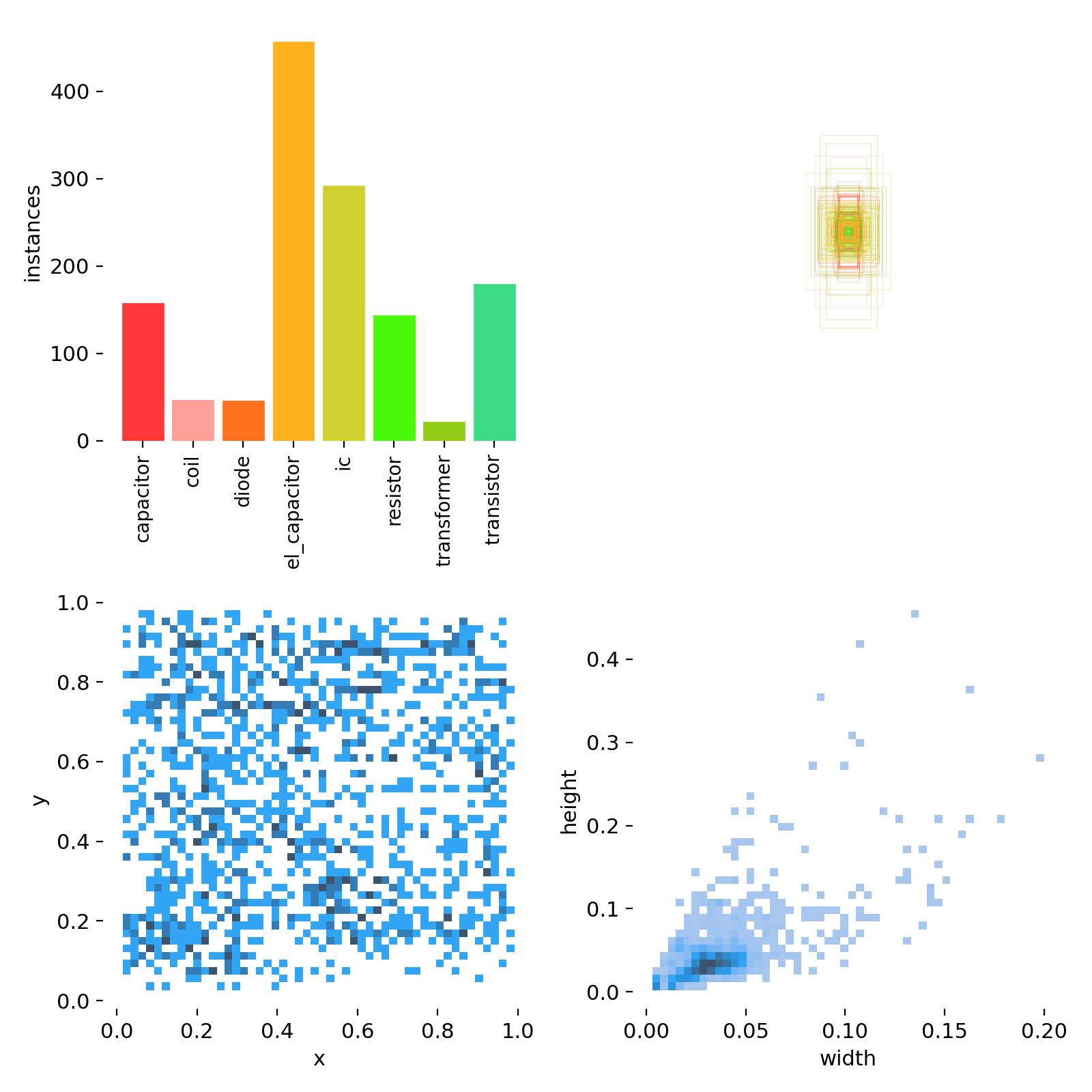}
\caption{Class-wise distribution of instances}
\label{fig:instance}
\end{figure}

Different recycling plants were visited during the data collection phase, and PCB boards were gathered from it which are further used to create the V-PCB dataset. A Raspberry Pi camera attached to an NVIDIA Jetson Nano, which offered high parallel processing capabilities for image analysis, was used to capture the dataset. The PCB raw images in the dataset have undergone preprocessing to improve image quality and lower noise. An online open-source tool called label studio \cite{labelstudio} was used to annotate the data, and eight classes of PCB components were chosen to train the deep learning model. 
The YOLOv5 architecture was used to train an object detection model during the model development and inference phases. The model was tested on the Jetson Nano environment after being trained on a Linux server with a fast GPU. In this study, the model demonstrated that it successfully identified maximum PCB components with accuracy and speed. Figure \ref{fig:instance} illustrates how many instances of each class are used during training and validation. Class instances with a higher number of instances perform well in the final inference of the model on Jetson Nano. For model generalization, data augmentation techniques has been applied to add more training data into V-PCBs dataset which contain more instances of those classes which are less in count. 
We trained the model using different batch sizes 4 and 8 respectively to analyses the behaviour and speed of model convergence. To evaluate the performance of the model, we used precision, recall, and mean average precision. Based on the equations \ref{eq:1},\ref{eq:2},\ref{eq:3}, model performance is evaluated on the basis of true positives, false positives, true negatives, and false negatives, both on the bounding box level and on the class level.

\begin{equation}
\label{eq:1}
Precision = \frac{TP}{TP + FP}
\end{equation}

\begin{equation}
\label{eq:2}
Recall = \frac{TP}{TP + FN}
\end{equation}

\begin{equation}
\label{eq:3}
mAP = \frac{1}{N}\sum_{i=1}^{N}AP_i
\end{equation}
Table \ref{tab:my_table} shows the results achieved on local V-PCB dataset. Based on it, the YOLOv5 model performed well and achieved the precision and recall of .80 and .56, respectively. 
\begin{table}[h!]
\caption{Performance metrics based on precision and recall}
    \centering
    \captionsetup{position=top, justification=centering} 
    \begin{tabular}{|c|c|c|c|c|c|c|}
    \hline
    \textbf{Val box loss} & \textbf{Val obj loss} & \textbf{Val cls loss} & \textbf{Precision} & \textbf{Recall} & \textbf{mAP}& \textbf{mAP-95}\\
    \hline
    0.04 & 0.05 & 0.02 & 0.82 & 0.55 & .56 & .27 \\
    \hline 
    \end{tabular}

\label{tab:my_table}
\end{table} 

\section{Conclusion}\label{sec5}
In this paper, we present ongoing work on improving the recycling of waste PCBs using deep learning and computer vision. It contributes to the concept of "virtual mines" in the circular economy by reclaiming valuable materials from end-of-life PCBs. Local V-PCB image dataset is prepared for component level recycling of waste PCBs. A deep learning model YOLOv5 is used for recognizing the electronic components on PCBs, and the results shown promising accuracy and speed. The proposed system could enhance the collection, disassembly, and recycling of PCBs, leading to greater environmental and economic benefits. In future, it is planned to introduce a hierarchical classification structure for recognising finer classes and possibly even read information from the PCB and from its components via OCR.

\input{sn-article.bbl}

\end{document}

%% file: sn-article.bbl

%% file: sn-article.bbl
\begin{thebibliography}{27}
\ifx \bisbn   \undefined \def \bisbn  #1{ISBN #1}\fi
\ifx \binits  \undefined \def \binits#1{#1}\fi
\ifx \bauthor  \undefined \def \bauthor#1{#1}\fi
\ifx \batitle  \undefined \def \batitle#1{#1}\fi
\ifx \bjtitle  \undefined \def \bjtitle#1{#1}\fi
\ifx \bvolume  \undefined \def \bvolume#1{\textbf{#1}}\fi
\ifx \byear  \undefined \def \byear#1{#1}\fi
\ifx \bissue  \undefined \def \bissue#1{#1}\fi
\ifx \bfpage  \undefined \def \bfpage#1{#1}\fi
\ifx \blpage  \undefined \def \blpage #1{#1}\fi
\ifx \burl  \undefined \def \burl#1{\textsf{#1}}\fi
\ifx \doiurl  \undefined \def \doiurl#1{\url{https://doi.org/#1}}\fi
\ifx \betal  \undefined \def \betal{\textit{et al.}}\fi
\ifx \binstitute  \undefined \def \binstitute#1{#1}\fi
\ifx \binstitutionaled  \undefined \def \binstitutionaled#1{#1}\fi
\ifx \bctitle  \undefined \def \bctitle#1{#1}\fi
\ifx \beditor  \undefined \def \beditor#1{#1}\fi
\ifx \bpublisher  \undefined \def \bpublisher#1{#1}\fi
\ifx \bbtitle  \undefined \def \bbtitle#1{#1}\fi
\ifx \bedition  \undefined \def \bedition#1{#1}\fi
\ifx \bseriesno  \undefined \def \bseriesno#1{#1}\fi
\ifx \blocation  \undefined \def \blocation#1{#1}\fi
\ifx \bsertitle  \undefined \def \bsertitle#1{#1}\fi
\ifx \bsnm \undefined \def \bsnm#1{#1}\fi
\ifx \bsuffix \undefined \def \bsuffix#1{#1}\fi
\ifx \bparticle \undefined \def \bparticle#1{#1}\fi
\ifx \barticle \undefined \def \barticle#1{#1}\fi
\bibcommenthead
\ifx \bconfdate \undefined \def \bconfdate #1{#1}\fi
\ifx \botherref \undefined \def \botherref #1{#1}\fi
\ifx \url \undefined \def \url#1{\textsf{#1}}\fi
\ifx \bchapter \undefined \def \bchapter#1{#1}\fi
\ifx \bbook \undefined \def \bbook#1{#1}\fi
\ifx \bcomment \undefined \def \bcomment#1{#1}\fi
\ifx \oauthor \undefined \def \oauthor#1{#1}\fi
\ifx \citeauthoryear \undefined \def \citeauthoryear#1{#1}\fi
\ifx \endbibitem  \undefined \def \endbibitem {}\fi
\ifx \bconflocation  \undefined \def \bconflocation#1{#1}\fi
\ifx \arxivurl  \undefined \def \arxivurl#1{\textsf{#1}}\fi
\csname PreBibitemsHook\endcsname

\bibitem[\protect\citeauthoryear{Forti et~al.}{2020}]{forti2020global}
\begin{botherref}
\oauthor{\bsnm{Forti}, \binits{V.}},
\oauthor{\bsnm{Balde}, \binits{C.P.}},
\oauthor{\bsnm{Kuehr}, \binits{R.}},
\oauthor{\bsnm{Bel}, \binits{G.}}:
The global e-waste monitor 2020: Quantities, flows and the circular economy
  potential
(2020)
\end{botherref}
\endbibitem

\bibitem[\protect\citeauthoryear{Zeng et~al.}{2018}]{zeng2018urban}
\begin{barticle}
\bauthor{\bsnm{Zeng}, \binits{X.}},
\bauthor{\bsnm{Mathews}, \binits{J.A.}},
\bauthor{\bsnm{Li}, \binits{J.}}:
\batitle{Urban mining of e-waste is becoming more cost-effective than virgin
  mining}.
\bjtitle{Environmental science \& technology}
\bvolume{52}(\bissue{8}),
\bfpage{4835}--\blpage{4841}
(\byear{2018})
\end{barticle}
\endbibitem

\bibitem[\protect\citeauthoryear{{European
  Commission}}{2020}]{european2020communication}
\begin{botherref}
\oauthor{\bsnm{{European Commission}}}:
Critical raw materials resilience: Charting a path towards greater security and
  sustainability.
Communication from the commission to the European Parliament, the Council, the
  European Economic and Social Committee and the committee of the regions
\textbf{474}
(2020)
\end{botherref}
\endbibitem

\bibitem[\protect\citeauthoryear{{European
  Commission}}{2011}]{european2011communication}
\begin{botherref}
\oauthor{\bsnm{{European Commission}}}:
Tackling the challenges in commodity markets and on raw materials.
Communication From The Commission To The European Parliament, The Council, The
  European Economic and Social Committee and The Committee of The Regions
\textbf{25}
(2011)
\end{botherref}
\endbibitem

\bibitem[\protect\citeauthoryear{{European
  Commission}}{2014}]{european2014communication}
\begin{botherref}
\oauthor{\bsnm{{European Commission}}}:
On the review of the list of critical raw materials for the eu and the
  implementation of the raw materials initiative.
Communication from the commission to the European Parliament, the Council, the
  European Economic and Social Committee and the committee of the regions
\textbf{297}
(2014)
\end{botherref}
\endbibitem

\bibitem[\protect\citeauthoryear{{European
  Commission}}{2017}]{european2017communication}
\begin{botherref}
\oauthor{\bsnm{{European Commission}}}:
On the 2017 list of critical raw materials for the eu.
Communication from the commission to the European Parliament, the Council, the
  European Economic and Social Committee and the committee of the regions
\textbf{490}
(2017)
\end{botherref}
\endbibitem

\bibitem[\protect\citeauthoryear{Cossu and Williams}{2015}]{mining2015urban}
\begin{barticle}
\bauthor{\bsnm{Cossu}, \binits{R.}},
\bauthor{\bsnm{Williams}, \binits{I.D.}}:
\batitle{Urban mining: Concepts, terminology, challenges}.
\bjtitle{Waste Management}
\bvolume{45},
\bfpage{1}--\blpage{3}
(\byear{2015})
\end{barticle}
\endbibitem

\bibitem[\protect\citeauthoryear{Xavier
  et~al.}{2021}]{xavier2021sustainability}
\begin{barticle}
\bauthor{\bsnm{Xavier}, \binits{L.H.}},
\bauthor{\bsnm{Giese}, \binits{E.C.}},
\bauthor{\bsnm{Ribeiro-Duthie}, \binits{A.C.}},
\bauthor{\bsnm{Lins}, \binits{F.A.F.}}:
\batitle{Sustainability and the circular economy: A theoretical approach
  focused on e-waste urban mining}.
\bjtitle{Resources Policy}
\bvolume{74},
\bfpage{101467}
(\byear{2021})
\end{barticle}
\endbibitem

\bibitem[\protect\citeauthoryear{Priya and Hait}{2017}]{priya2017comparative}
\begin{barticle}
\bauthor{\bsnm{Priya}, \binits{A.}},
\bauthor{\bsnm{Hait}, \binits{S.}}:
\batitle{Comparative assessment of metallurgical recovery of metals from
  electronic waste with special emphasis on bioleaching}.
\bjtitle{Environmental Science and Pollution Research}
\bvolume{24},
\bfpage{6989}--\blpage{7008}
(\byear{2017})
\end{barticle}
\endbibitem

\bibitem[\protect\citeauthoryear{Xiang et~al.}{2010}]{xiang2010bioleaching}
\begin{barticle}
\bauthor{\bsnm{Xiang}, \binits{Y.}},
\bauthor{\bsnm{Wu}, \binits{P.}},
\bauthor{\bsnm{Zhu}, \binits{N.}},
\bauthor{\bsnm{Zhang}, \binits{T.}},
\bauthor{\bsnm{Liu}, \binits{W.}},
\bauthor{\bsnm{Wu}, \binits{J.}},
\bauthor{\bsnm{Li}, \binits{P.}}:
\batitle{Bioleaching of copper from waste printed circuit boards by bacterial
  consortium enriched from acid mine drainage}.
\bjtitle{Journal of hazardous materials}
\bvolume{184}(\bissue{1-3}),
\bfpage{812}--\blpage{818}
(\byear{2010})
\end{barticle}
\endbibitem

\bibitem[\protect\citeauthoryear{Nguyen et~al.}{2021}]{nguyen2021bioleaching}
\begin{barticle}
\bauthor{\bsnm{Nguyen}, \binits{T.H.}},
\bauthor{\bsnm{Won}, \binits{S.}},
\bauthor{\bsnm{Ha}, \binits{M.-G.}},
\bauthor{\bsnm{Nguyen}, \binits{D.D.}},
\bauthor{\bsnm{Kang}, \binits{H.Y.}}:
\batitle{Bioleaching for environmental remediation of toxic metals and
  metalloids: A review on soils, sediments, and mine tailings}.
\bjtitle{Chemosphere}
\bvolume{282},
\bfpage{131108}
(\byear{2021})
\end{barticle}
\endbibitem

\bibitem[\protect\citeauthoryear{Yaashikaa et~al.}{2022}]{yaashikaa2022review}
\begin{barticle}
\bauthor{\bsnm{Yaashikaa}, \binits{P.}},
\bauthor{\bsnm{Priyanka}, \binits{B.}},
\bauthor{\bsnm{Kumar}, \binits{P.S.}},
\bauthor{\bsnm{Karishma}, \binits{S.}},
\bauthor{\bsnm{Jeevanantham}, \binits{S.}},
\bauthor{\bsnm{Indraganti}, \binits{S.}}:
\batitle{A review on recent advancements in recovery of valuable and toxic
  metals from e-waste using bioleaching approach}.
\bjtitle{Chemosphere}
\bvolume{287},
\bfpage{132230}
(\byear{2022})
\end{barticle}
\endbibitem

\bibitem[\protect\citeauthoryear{Cabri et~al.}{2022}]{cabri2022recovering}
\begin{bchapter}
\bauthor{\bsnm{Cabri}, \binits{A.}},
\bauthor{\bsnm{Masulli}, \binits{F.}},
\bauthor{\bsnm{Rovetta}, \binits{S.}},
\bauthor{\bsnm{Mohsin}, \binits{M.}}, \betal:
\bctitle{Recovering critical raw materials from weee using artificial
  intelligence}.
In: \bbtitle{Proceedings of the 21st International Conference on Modelling and
  Applied Simulation MAS},
pp. \bfpage{1}--\blpage{5}
(\byear{2022}).
\bcomment{[sl]}
\end{bchapter}
\endbibitem

\bibitem[\protect\citeauthoryear{Kamble et~al.}{2022}]{kamble2022digital}
\begin{barticle}
\bauthor{\bsnm{Kamble}, \binits{S.S.}},
\bauthor{\bsnm{Gunasekaran}, \binits{A.}},
\bauthor{\bsnm{Parekh}, \binits{H.}},
\bauthor{\bsnm{Mani}, \binits{V.}},
\bauthor{\bsnm{Belhadi}, \binits{A.}},
\bauthor{\bsnm{Sharma}, \binits{R.}}:
\batitle{Digital twin for sustainable manufacturing supply chains: Current
  trends, future perspectives, and an implementation framework}.
\bjtitle{Technological Forecasting and Social Change}
\bvolume{176},
\bfpage{121448}
(\byear{2022})
\end{barticle}
\endbibitem

\bibitem[\protect\citeauthoryear{Mohsin et~al.}{2021}]{mohsin2021automatic}
\begin{bchapter}
\bauthor{\bsnm{Mohsin}, \binits{M.}},
\bauthor{\bsnm{Shaukat}, \binits{A.}},
\bauthor{\bsnm{Akram}, \binits{U.}},
\bauthor{\bsnm{Zarrar}, \binits{M.K.}}:
\bctitle{Automatic prostate cancer grading using deep architectures}.
In: \bbtitle{2021 IEEE/ACS 18th International Conference on Computer Systems
  and Applications (AICCSA)},
pp. \bfpage{1}--\blpage{8}
(\byear{2021}).
\bcomment{IEEE}
\end{bchapter}
\endbibitem

\bibitem[\protect\citeauthoryear{Adibhatla
  et~al.}{2018}]{adibhatla2018detecting}
\begin{bchapter}
\bauthor{\bsnm{Adibhatla}, \binits{V.A.}},
\bauthor{\bsnm{Shieh}, \binits{J.-S.}},
\bauthor{\bsnm{Abbod}, \binits{M.F.}},
\bauthor{\bsnm{Chih}, \binits{H.-C.}},
\bauthor{\bsnm{Hsu}, \binits{C.-C.}},
\bauthor{\bsnm{Cheng}, \binits{J.}}:
\bctitle{Detecting defects in pcb using deep learning via convolution neural
  networks}.
In: \bbtitle{2018 13th International Microsystems, Packaging, Assembly and
  Circuits Technology Conference (IMPACT)},
pp. \bfpage{202}--\blpage{205}
(\byear{2018}).
\bcomment{IEEE}
\end{bchapter}
\endbibitem

\bibitem[\protect\citeauthoryear{Lu et~al.}{2020}]{lu2020fics}
\begin{botherref}
\oauthor{\bsnm{Lu}, \binits{H.}},
\oauthor{\bsnm{Mehta}, \binits{D.}},
\oauthor{\bsnm{Paradis}, \binits{O.}},
\oauthor{\bsnm{Asadizanjani}, \binits{N.}},
\oauthor{\bsnm{Tehranipoor}, \binits{M.}},
\oauthor{\bsnm{Woodard}, \binits{D.L.}}:
Fics-pcb: A multi-modal image dataset for automated printed circuit board
  visual inspection.
Cryptology ePrint Archive
(2020)
\end{botherref}
\endbibitem

\bibitem[\protect\citeauthoryear{Li et~al.}{2013}]{li2013smd}
\begin{bchapter}
\bauthor{\bsnm{Li}, \binits{W.}},
\bauthor{\bsnm{Esders}, \binits{B.}},
\bauthor{\bsnm{Breier}, \binits{M.}}:
\bctitle{Smd segmentation for automated pcb recycling}.
In: \bbtitle{2013 11th IEEE International Conference on Industrial Informatics
  (INDIN)},
pp. \bfpage{65}--\blpage{70}
(\byear{2013}).
\bcomment{IEEE}
\end{bchapter}
\endbibitem

\bibitem[\protect\citeauthoryear{Li et~al.}{2016}]{li2016localizing}
\begin{bchapter}
\bauthor{\bsnm{Li}, \binits{W.}},
\bauthor{\bsnm{Jiang}, \binits{C.}},
\bauthor{\bsnm{Breier}, \binits{M.}},
\bauthor{\bsnm{Merhof}, \binits{D.}}:
\bctitle{Localizing components on printed circuit boards using 2d information}.
In: \bbtitle{2016 IEEE International Conference on Industrial Technology
  (ICIT)},
pp. \bfpage{769}--\blpage{774}
(\byear{2016}).
\bcomment{IEEE}
\end{bchapter}
\endbibitem

\bibitem[\protect\citeauthoryear{Li et~al.}{2014}]{li2014text}
\begin{bchapter}
\bauthor{\bsnm{Li}, \binits{W.}},
\bauthor{\bsnm{Neullens}, \binits{S.}},
\bauthor{\bsnm{Breier}, \binits{M.}},
\bauthor{\bsnm{Bosling}, \binits{M.}},
\bauthor{\bsnm{Pretz}, \binits{T.}},
\bauthor{\bsnm{Merhof}, \binits{D.}}:
\bctitle{Text recognition for information retrieval in images of printed
  circuit boards}.
In: \bbtitle{IECON 2014-40th Annual Conference of the IEEE Industrial
  Electronics Society},
pp. \bfpage{3487}--\blpage{3493}
(\byear{2014}).
\bcomment{IEEE}
\end{bchapter}
\endbibitem

\bibitem[\protect\citeauthoryear{Xu et~al.}{2020}]{xu2020n}
\begin{barticle}
\bauthor{\bsnm{Xu}, \binits{Y.}},
\bauthor{\bsnm{Yang}, \binits{G.}},
\bauthor{\bsnm{Luo}, \binits{J.}},
\bauthor{\bsnm{He}, \binits{J.}}:
\batitle{n electronic component recognition algorithm based on deep learning
  with a faster squeezenet}.
\bjtitle{Mathematical Problems in Engineering}
\bvolume{2020},
\bfpage{1}--\blpage{11}
(\byear{2020})
\end{barticle}
\endbibitem

\bibitem[\protect\citeauthoryear{Huang et~al.}{2016}]{huang2016learning}
\begin{bchapter}
\bauthor{\bsnm{Huang}, \binits{C.}},
\bauthor{\bsnm{Li}, \binits{Y.}},
\bauthor{\bsnm{Loy}, \binits{C.C.}},
\bauthor{\bsnm{Tang}, \binits{X.}}:
\bctitle{Learning deep representation for imbalanced classification}.
In: \bbtitle{Proceedings of the IEEE Conference on Computer Vision and Pattern
  Recognition},
pp. \bfpage{5375}--\blpage{5384}
(\byear{2016})
\end{bchapter}
\endbibitem

\bibitem[\protect\citeauthoryear{Japkowicz and
  Stephen}{2002}]{japkowicz2002class}
\begin{barticle}
\bauthor{\bsnm{Japkowicz}, \binits{N.}},
\bauthor{\bsnm{Stephen}, \binits{S.}}:
\batitle{The class imbalance problem: A systematic study}.
\bjtitle{Intelligent data analysis}
\bvolume{6}(\bissue{5}),
\bfpage{429}--\blpage{449}
(\byear{2002})
\end{barticle}
\endbibitem

\bibitem[\protect\citeauthoryear{Fan et~al.}{2019}]{fan2019birnet}
\begin{barticle}
\bauthor{\bsnm{Fan}, \binits{J.}},
\bauthor{\bsnm{Cao}, \binits{X.}},
\bauthor{\bsnm{Yap}, \binits{P.-T.}},
\bauthor{\bsnm{Shen}, \binits{D.}}:
\batitle{Birnet: Brain image registration using dual-supervised fully
  convolutional networks}.
\bjtitle{Medical image analysis}
\bvolume{54},
\bfpage{193}--\blpage{206}
(\byear{2019})
\end{barticle}
\endbibitem

\bibitem[\protect\citeauthoryear{Tan and Le}{2019}]{tan2019efficientnet}
\begin{bchapter}
\bauthor{\bsnm{Tan}, \binits{M.}},
\bauthor{\bsnm{Le}, \binits{Q.}}:
\bctitle{Efficientnet: Rethinking model scaling for convolutional neural
  networks}.
In: \bbtitle{International Conference on Machine Learning},
pp. \bfpage{6105}--\blpage{6114}
(\byear{2019}).
\bcomment{PMLR}
\end{bchapter}
\endbibitem

\bibitem[\protect\citeauthoryear{OpenAI}{2022}]{labelstudio}
\begin{botherref}
\oauthor{\bsnm{OpenAI}}:
Label Studio v1.6.0
(2022).
\url{https://labelstud.io}
Accessed 05.12.2022
\end{botherref}
\endbibitem

\bibitem[\protect\citeauthoryear{Jocher et~al.}{2022}]{jocher2022ultralytics}
\begin{botherref}
\oauthor{\bsnm{Jocher}, \binits{G.}},
\oauthor{\bsnm{Chaurasia}, \binits{A.}},
\oauthor{\bsnm{Stoken}, \binits{A.}},
\oauthor{\bsnm{Borovec}, \binits{J.}},
\oauthor{\bsnm{Kwon}, \binits{Y.}},
\oauthor{\bsnm{Michael}, \binits{K.}},
\oauthor{\bsnm{Fang}, \binits{J.}},
\oauthor{\bsnm{Yifu}, \binits{Z.}},
\oauthor{\bsnm{Wong}, \binits{C.}},
\oauthor{\bsnm{Montes}, \binits{D.}}, et al.:
ultralytics/yolov5: V7. 0-yolov5 sota realtime instance segmentation.
Zenodo
(2022)
\end{botherref}
\endbibitem

\end{thebibliography}
